\documentclass[11pt,journal,compsoc]{IEEEtran}
\usepackage{times}
\usepackage{epsfig}
\usepackage{caption}
\usepackage{subfigure}
\usepackage{graphicx, float, subfigure}
\usepackage{amsmath}
\usepackage{multirow}
\usepackage{color}
\usepackage{amssymb}
\usepackage{booktabs}
\usepackage{tabularx}
\usepackage{makecell}
\usepackage{setspace}
\usepackage[pagebackref=true,breaklinks=true,letterpaper=true,colorlinks,bookmarks=false]{hyperref}
\usepackage{float}
\usepackage{aliascnt}
\usepackage{enumerate}
\ifCLASSOPTIONcompsoc
  \usepackage[nocompress]{cite}
\else
  \usepackage{cite}
\fi

\ifCLASSINFOpdf
\else
\fi

\begin{document}
\title{SPRNet: Single Pixel Reconstruction for \\ One-stage Instance Segmentation}


\author{Jun~Yu,~\IEEEmembership{Member,~IEEE,}
        Jinghan~Yao,
        Jian~Zhang,
        Zhou~Yu,
        and~Dacheng~Tao,~\IEEEmembership{Fellow,~IEEE}
\thanks{
This work was supported by the National Natural Science Foundation of
China under Grant 61836002, Grant 61622205, by the Zhejiang Provincial Natural Science Foundation of China under Grant LY17F020009, and by the Australian Research Council Projects under Grant FL-170100117,
Grant DP-180103424, and Grant IH-180100002.
(Jun Yu and Jinghan Yao contributed equally to this work.)
(Corresponding author: Jian Zhang).}

\thanks{J. Yu is with the School of Computer Science, Hangzhou Dianzi University,
Hangzhou 310018, China (email: yujun@hdu.edu.cn).}
\thanks{J. Yao is with the School of Computer Science, Hangzhou Dianzi University,
Hangzhou 310018, China (email: yjhmitweb@gmail.com).}
\thanks{J. Zhang is with the School of Science and Technology, Zhejiang
International Studies University, Hangzhou 310023, China (email:
jzhang@zisu.edu.cn). Corresponding author.}
\thanks{Z. Yu is with the School of Computer Science, Hangzhou Dianzi University,
Hangzhou 310018, China (email: yuz@hdu.edu.cn ).}
\thanks{D. Tao is with the UBTECH Sydney Artificial Intelligence Centre and the School of Computer Science, in the Faculty of Engineering and Information Technologies, at the University of Sydney, 6 Cleveland St, Darlington, NSW 2008, Australia (email: dacheng.tao@sydney.edu.au).}
}

\markboth{A Submission to IEEE Transactions on Cybernetics}%
{Shell \MakeLowercase{\textit{et al.}}: Bare Demo of IEEEtran.cls for Computer Society Journals}

\clubpenalty=10000
\widowpenalty = 10000
\hyphenpenalty=5000
\tolerance=1000

\IEEEtitleabstractindextext{%
\begin{abstract}
  Object instance segmentation is one of the most fundamental but challenging tasks in computer vision, and it requires the pixel-level image understanding. Most existing approaches address this problem by adding a mask prediction branch to a two-stage object detector with the Region Proposal Network (RPN). Although producing good segmentation results, the efficiency of these two-stage approaches is far from satisfactory, restricting their applicability in practice. In this paper, we propose a one-stage framework, SPRNet, which performs efficient instance segmentation by introducing a single pixel reconstruction (SPR) branch to off-the-shelf one-stage detectors. The added SPR branch reconstructs the pixel-level mask from every single pixel in the convolution feature map directly. Using the same ResNet-50 backbone, SPRNet achieves comparable mask AP to Mask R-CNN at a higher inference speed, and gains all-round improvements on box AP at every scale comparing \textcolor{black}{with} RetinaNet.
\end{abstract}

}

\maketitle

\IEEEdisplaynontitleabstractindextext
\IEEEpeerreviewmaketitle

\IEEEraisesectionheading{\section{Introduction}\label{sec:introduction}}

\IEEEPARstart{R}{apid} developments in object detection \cite{zhou2019} and semantic segmentation \cite{wang2019} have made it possible to accurately \textcolor{black}{understand} region or pixel-level semantics \textcolor{black}{of images}. An increasing number of researchers focus more on real-time capability of models \cite{Shen2018Multiobject}. \textcolor{black}{Rather than achieving object detection and semantic segmentation independently, integrating both of them into one framework is more preferable but a more challenging instance segmentation task in which not only the class label predictions of all the objects but also their pixel-level predictions separating the objects from the background are needed.} Optimally addressing this task would greatly benefit many applications including autonomous driving, robotics and video surveillance.

Most successful instance segmentation approaches are derived from object detection models \textcolor{black}{by} adding a new branch on top of regional features from Region Proposal Network (RPN) \cite{ren2015faster} to predict \textcolor{black}{the objects'} corresponding masks \cite{he2017mask,Yi2017Fully,liu2018path}. \textcolor{black}{This model can be referred to as the two-stage instance segmentation model whose efficiency is often influenced by the detection-based scheme.} \textcolor{black}{A well-known example is the Mask R-CNN \cite{he2017mask} framework built on the Faster R-CNN \cite{ren2015faster} backbone, in which the efficiency is limited by the two-stage mechanism with a heavy head and multiple Regions of Interest (RoIs) \cite{ren2015faster}.}

\textcolor{black}{Apart from the two-stage scheme, one-stage scheme has been applied to object detection and the state-of-the-art detectors have reported similar accuracy to two-stage detector with faster speed.}
With carefully designed architectures \cite{liu2016ssd,Redmon2017YOLO9000,lin2017feature} and loss functions \cite{lin2018focal}, one-stage detectors are able to run in real-time. This gives rise to a natural question: \emph{can we design a one-stage instance segmentation model that enjoys the dual benefits of efficiency and accuracy?}

\begin{figure}
	\centering
	\includegraphics[width=0.45\textwidth]{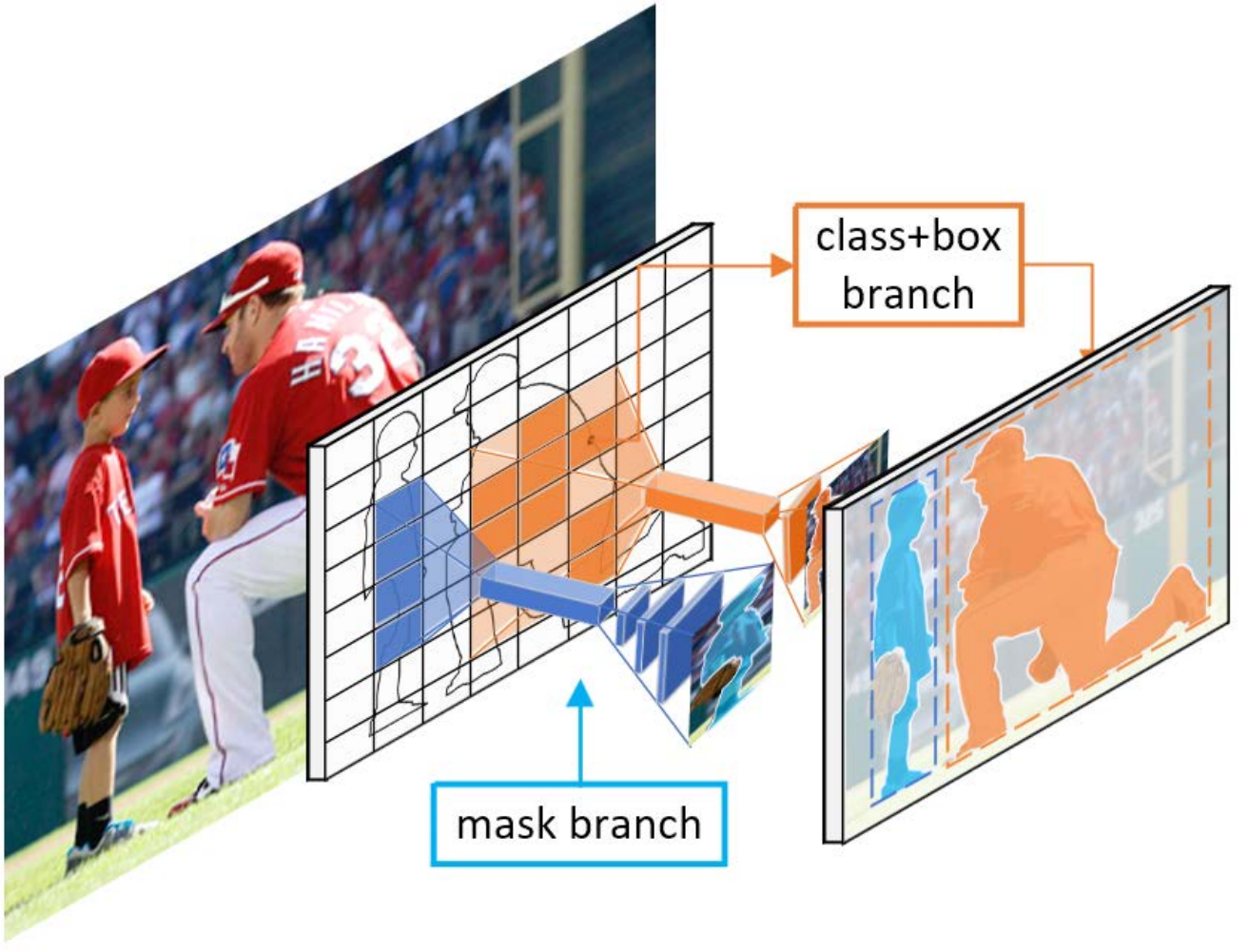}
	\caption{The SPRNet framework for instance segmentation. Classification, regression and mask branches are processed in parallel. We generate each instance mask from a single pixel, \textcolor{black}{and resize the mask to fit the corresponding box to get the final instance-level prediction}.}
	\label{fig:1}
\end{figure}

The main problem in one-stage instance segmentation is how to distinguish instances between and within classes simultaneously from a convolution feature map without the help of pre-generated RoIs. To address this problem, we propose a new framework named as the \textcolor{black}{Single Pixel Reconstruction Net (SPRNet) described in Figure \ref{fig:1}}. By using only a pixel to reconstruct the mask of an instance, \textcolor{black}{our method outperforms R-CNN based frameworks where the prediction of instance mask starts from each RoI in that we can predict the instance mask in a faster way with less memory consumption and allow more instances to be predicted simultaneously on an image.}
We use a group of convolutions with various dilations to gather enough information into a single pixel. To construct the mask of one instance, we sample a pixel and then use consecutive deconvolutions to construct a 32$\times$32 score map denoting the final mask prediction.

\textcolor{black}{We also make modifications to the backbone part in our model such that we can enhance the information carried by each pixel to facilitate the implementation of the SPRNet.} \textcolor{black}{Specifically,}
we borrow the principle from Feature Pyramid Network (FPN) \cite{lin2017feature}, where features from different levels are added to construct a {bottom-up and top-down} path. In their implementation, a naive element-wise summation is used to fuse multi-level features.
\textcolor{black}{However, the FPN framework has two problems.}
\textcolor{black}{First, the summation operation may introduce unexpected information flow in feature learning in that} 
severe spatial shift from higher levels could damage already well preserved spatial information \textcolor{black}{in} lower ones.
\textcolor{black}{Second}, since the derivative of summation is always a constant, summation will cause gradient propagation between \textcolor{black}{different levels}, implicitly weakening the effect of the fused features. \textcolor{black}{This is} because the initiative of designing FPN is to detect objects in different scales on different levels, which means that gradient from one level should not easily interfere \textcolor{black}{that of another}. Therefore, we propose an improved Gate-FPN (GFPN) by explicitly introducing a simple gating mechanism before feature fusion. This step improves the quality of feature fusion and smartly restrict gradient propagation between different levels, leading to better detection and segmentation.

\par \textcolor{black}{The proposed} SPRNet is of an encoder-decoder structure. In encoding part, following common backbones, Gate-FPN enhances semantic and spatial information carried by each pixel. In decoding part, each pixel is an instance carrier to generate instance mask, consecutive deconvolutions are applied on them to get the final predictions. To summarize, SPRNet extends the state-of-the-art one-stage detector RetinaNet by adding a parallel branch for predicting an object mask. Note that SPRNet is a general framework, so the backbone network can be replaced with other one-stage detectors without damaging the integrity and feasibility of instance mask generation. Our main contributions are as follows:
\begin{enumerate}[(i)]
\item To the best of our knowledge, we propose a one-stage instance segmentation model for the first time. SPRNet achieves comparable performance in terms of mask AP with Mask R-CNN, while delivering faster speed at the inference;
\item By introducing the Gate-FPN architecture, we bring all-round improvements on AP in detection and segmentation performance.
\end{enumerate}

\textcolor{black}{The rest part of the paper is organized as follows. Section \ref{review} reviews the related work. Section \ref{SPRNet} introduces the proposed SPRNet model with Gate-FPN in detail. Experimental results are demonstrated and analyzed in Section \ref{exp:seg} through Section \ref{add:exp} and Section \ref{conc} concludes this paper.}

\section{Related work}
\label{review}
\vspace{8pt}

\noindent \textbf{Object Detection:} Prevalent object detection approaches can be categorized into two classes: the two-stage framework based on region proposals \cite{girshick2015fast,ren2015faster}, or the one-stage framework based on convolutional feature maps \cite{redmon2016you,liu2016ssd,lin2018focal}. As pioneered in the R-CNN work \cite{ren2015faster}, the first stage \textcolor{black}{generates} a set of candidate region proposals to recall as much objects as possible, and the second stage \textcolor{black}{uses} a deep \textcolor{black}{network} to classify the proposals. R-CNN successfully underpinned many follow-up improvements like Faster R-CNN \cite{ren2015faster} and R-FCN \cite{DBLP:journals/corr/DaiLHS16}, which are the current leading frameworks for object detection. An inevitable procedure of the two-stage methods is the per-proposal prediction, \textcolor{black}{which becomes a speed bottleneck of these methods in case of large number of proposals}. In contrast, the one-stage methods do not introduce the region proposals thus \textcolor{black}{have high computational efficiency}. \textcolor{black}{Commonly adopted object detectors include OverFeat \cite{Sermanet2013OverFeat}, SSD \cite{liu2016ssd}, YOLO \cite{redmon2016you}, DSSD \cite{Fu2017DSSD} and RetinaNet \cite{lin2018focal}.} OverFeat \textcolor{black}{is} one of the first modern one-stage object detectors based on deep networks. SSD \cite{liu2016ssd} and YOLO \cite{redmon2016you} are carefully designed \textcolor{black}{to speedup the implementation} at the \textcolor{black}{cost} of \textcolor{black}{degradation in} accuracy. Recently, DSSD \cite{Fu2017DSSD} and RetinaNet \cite{lin2018focal} have renewed interest in one-stage methods, achieving impressive accuracy that rivals that of two-stage detectors, and also running at much higher speeds.

\vspace{6pt}

\noindent \textbf{Instance Segmentation:} Instance segmentation requires a pixel-level prediction between and within classes \cite{hariharan2014simultaneous}. \textcolor{black}{Current instance segmentation methods can be categorized into two classes, i.e., detection-dependent and detection-free method.}

Combining the object detection task with the semantic segmentation task results in a more challenging instance segmentation task. DeepMask \cite{pinheiro2015learning} and subsequent works first generate candidate segment proposals, and then classify them by Fast R-CNN \cite{girshick2015fast}. As the segmentation stage is time-consuming, these methods are usually slow. Fully Convolutional Instance Segmentation (FCIS) \cite{Yi2017Fully} \textcolor{black}{embeds} the segment proposal stage into an object detection framework. However, FCIS \textcolor{black}{produces} systematic errors for overlapping instances and creates spurious edges, demonstrating that it is \textcolor{black}{vulnerable to} the inherent difficulties in segmenting instances. Mask R-CNN extends Faster R-CNN by adding a branch for predicting segmentation masks on each Region of Interest (RoI) \cite{he2017mask}. All the above instance segmentation approaches are two-stage \textcolor{black}{methods} that \textcolor{black}{need} to generate region (or segment) proposals first, which \textcolor{black}{limits} their running speed. \textcolor{black}{One-stage object detectors work well because they only need to predict rectangular boxes for describing objects}. However, generating promising instance-level segmentation masks is beyond the capability of existing one-stage detectors.

Detection-free methods are more \textcolor{black}{straightforward}. Since common semantic segmentation could be well solved, which is relatively easier than classic proposal models, this kind of methods extend common semantic segmentation models like \cite{chen2018deeplab,long2015fully} by \textcolor{black}{exploiting} hand-designed clustering algorithms. Once all pixels in an image are classified, clustering will be used to group them together for different object instances.

\vspace{6pt}

\noindent \textcolor{black}{\textbf{Multi-Scale Feature Learning:}} In both the detection and segmentation tasks, the problem of detecting small objects is an open problem. On the bottom layers of a deep \textcolor{black}{network}, objects have rich spatial information but \textcolor{black}{lack} clear semantic; while on the top layers the situation \textcolor{black}{turns} around. To tackle this problem, Feature Pyramid Network (FPN) was proposed to aggregate multi-level features \cite{wangx2019} in a pyramidal hierarchy. Beyond the traditional \emph{bottom-up} pathway \textcolor{black}{in} the deep networks,  FPN additionally \textcolor{black}{establishes} a \emph{top-down} paths to deliver the rich semantic information of top layers to the bottom layers \cite{lin2017feature}.
In their implementation, each \textcolor{black}{higher} layer feature \textcolor{black}{map} is 2$\times$ up-sampled and then merged with the \textcolor{black}{and lower} level feature \textcolor{black}{map} using element-wise summation. By doing so, the semantic information of top layers can be passed to the bottom layers successfully. However, their inaccurate spatial information is also passed to the bottom layers at the same time, which \textcolor{black}{\emph{counteracts}} the effectiveness to some extent. 
\begin{figure*}
	\centering
	\includegraphics[width=1.0\textwidth]{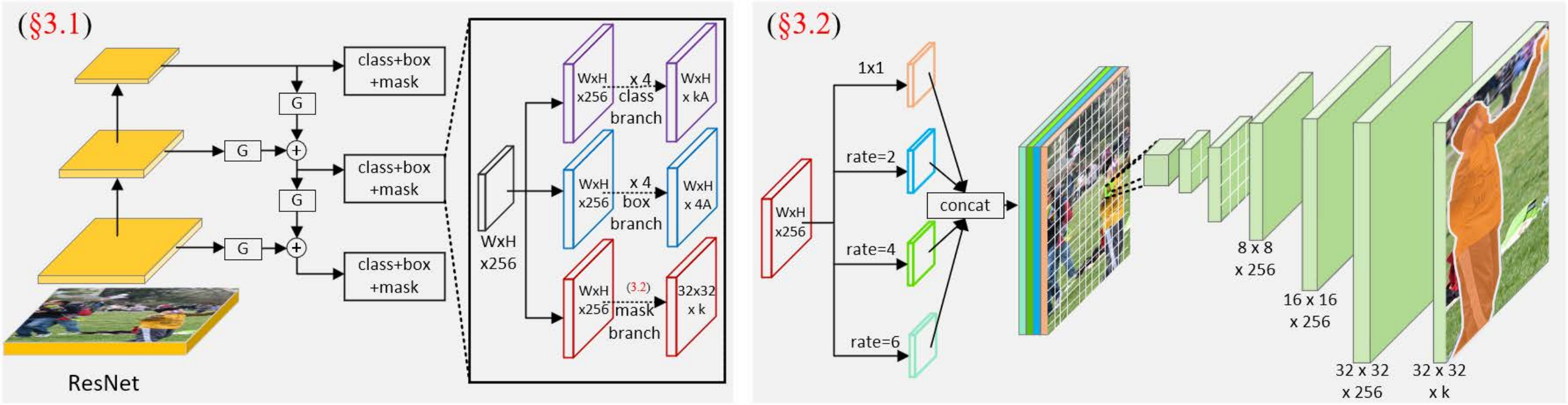}
	\caption{\textbf{SPRNet:}  \textit{A} denotes the number of anchors, \textit{k} denotes the number of classes. ({\color{black} 3.1}) \textcolor{black}{is the detailed architecture. The backbone network is ResNet-50, where 4$\times$, 8$\times$, 16$\times$, 32$\times$ and 64$\times$ downsampled feature maps are used to construct the FPN. The 32$\times$ downsampled feature map is the pooling of the 16$\times$ downsampled one, the 64$\times$ downsampled feature map is the convolutional result on 32$\times$ downsampled feature map with stride 2, and \emph{G} denotes the gate mechanism}. \textcolor{black}{The gated FPN is followed by three parallel branches for predicting class, box and mask respectively}. ({\color{black} 3.2}) \textcolor{black}{represents the spreading of the mask branch that includes} multi-scale fusion, channel-wise concatenation, positive pixel sampling and consecutive deconvolutions for instance mask generation. Note that each deconvolution will result in a 2$\times$ wider feature map, so that 5 deconvolution layer will decode a 1$\times$1 pixel into a 32$\times$32 mask.}
	\label{fig:2}
\end{figure*}
\noindent

\section{Single Pixel Reconstruction \textcolor{black}{Network}}
\label{SPRNet}
\vspace{8pt}

SPRNet \textcolor{black}{shares similarity with} detection\textcolor{black}{-}based instance segmentation frameworks. \textcolor{black}{However,} unlike Mask R-CNN or any other existing two-stage methods, SPRNet is free of RoI Pooling, RoI Align or any similar mechanisms. The whole instance segmentation process in SPRNet is \textcolor{black}{seamlessly integrated}, SPRNet \textcolor{black}{can therefore be viewed as} a detection-based one-stage framework for instance segmentation.
\par To achieve the goal of one-stage instance segmentation, there are two core problems to be well resolved: 1) how to \emph{encode} sufficient spatial and semantic information into each single pixel \textcolor{black}{of the feature maps;}
2) how to \emph{decode} the instance mask with respect to a single pixel from feature \textcolor{black}{maps}.  \textcolor{black}{To address the two problems}, we first introduce the network architecture of SPRNet \textcolor{black}{that} is inspired by RetinaNet \cite{lin2018focal} and motivated by the \emph{deconvolution} introduced in the semantic segmentation models \cite{chen2018deeplab, long2015fully}, we \textcolor{black}{subsequently} propose a mask branch that consists of a cascade of deconvolution layers to reconstruct complete mask from a single pixel. \textcolor{black}{The detailed architecture of SPRNet is demonstrated in Figure \ref{fig:2} where the left panel corresponds to problem 1) and the right panel corresponds to problem 2) }.

\subsection{Network Architecture}
SPRNet uses ResNet or any other feature extractors with \textcolor{black}{deep network} structures as the backbone. \textcolor{black}{We tackle the first problem, i.e. spatial and semantic information encoding, through multi-scale feature learning according to which} the feature maps at \textcolor{black}{multiple} scales, \textcolor{black}{e.g. those produced by the third, fourth and fifth convolutional layers (C3, C4 and C5)}, are applied for multi-scale detections.
The network architecture of SPRNet is \textcolor{black}{somewhat} similar \textcolor{black}{to} RetinaNet \cite{lin2018focal} \textcolor{black}{that} is a one-stage object detector with ResNet and FPN as its backbone. \textcolor{black}{However,} our SPRNet \textcolor{black}{is modified to overcome two primary problems of RetinaNet} \textcolor{black}{such that it works better in detection and is competent for instance segmentation}.

\vspace{4pt}

First, \textcolor{black}{higher level feature maps are up-sampled and directly added to \textcolor{black}{the lower level} ones to form feature maps of FPN. The basic concern is to take advantage of both the spatial information contained in the lower level features and the semantic information contained in the higher level features to achieve multi-scale object detection. Typically, a lower level feature map is twice the length and width as the higher level feature map. However, in ResNet \cite{xie2017aggregated, he2016deep}, there could be dozens of layers between such pair of higher and lower level feature maps, which means the structure information contained in the higher level feature maps are highly compact. Therefore, up-sampling the higher level feature maps will lead to ambiguity of object locations and we name this phenomena as spatial shift, which will impose unexpected influence when adding the up-sampled feature maps to lower level ones. }

\vspace{4pt}
\par
\begin{figure}
	\centering
	\includegraphics[width=0.45\textwidth]{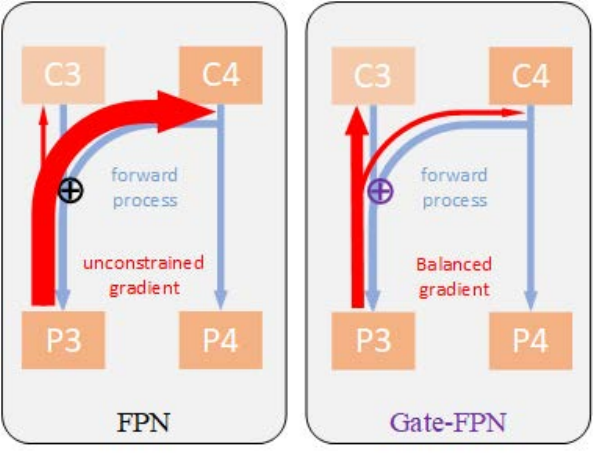}
	\caption{Unconstrained gradient will result in unexpected and unstable training process \textcolor{black}{of FPN}. \textcolor{black}{As is shown in the FPN panel},the gradients \textcolor{black}{propagated} from P3 \textcolor{black}{(the level-3 feature map of FPN)} to C3 \textcolor{black}{(the level-3 feature map of the backbone network)} is much larger than \textcolor{black}{those propagated} to C4, such that C3 only has subtle influence on P3 compared \textcolor{black}{with} C4 \textcolor{black}{in the procedure of optimization}. \textcolor{black}{This incurs serious gradient bias in that} C3 has not been well exploited \textcolor{black}{during optimization}. While in \textcolor{black}{the Gate-FPN panel}, gradients \textcolor{black}{passing to} both sides will be tuned first such that \textcolor{black}{C3 is the dominant contributor}. \textcolor{black}{Note that C3 is supposed to be the primary contributor because we do not encourage the gradient propagation from P3 to C4 to avoid spatial shift introduced by C4 and force P3 to focus on the objects with specific size.}}
	\label{fig:31}
\end{figure}

 Second, \textcolor{black}{adding operation may cause unexpected gradient propagation.} 
The derivative of \textcolor{black}{`a+b' with respect to `a' is constant `b' and vice versa.} If we perform `+' between two levels of features in the backbone network, during gradient back-propagation \cite{lecun1990handwritten}, the loss from the lower \textcolor{black}{level summed} features will directly \textcolor{black}{be passed to higher level} features (please refer to formulas (4) through (10)).
\textcolor{black}{We nevertheless hope different level of feature pyramid can concentrate on detecting objects with corresponding scale, therefore we do not encourage the loss of lower level summed features to interfere with higher level feature maps that introduce spatial shift into the lower level. A typical situation is shown in Figure \ref{fig:31}, where gradients propagated from P3 \textcolor{black}{(the level-3 feature map of FPN)} to C4 \textcolor{black}{(the level-4 feature map of the backbone network)} have completely overwhelmed that to C3 such that C4 plays dominant role in the learning of P3 and this is actually unwanted.}

\begin{figure}
	\centering
	\includegraphics[width=0.45\textwidth]{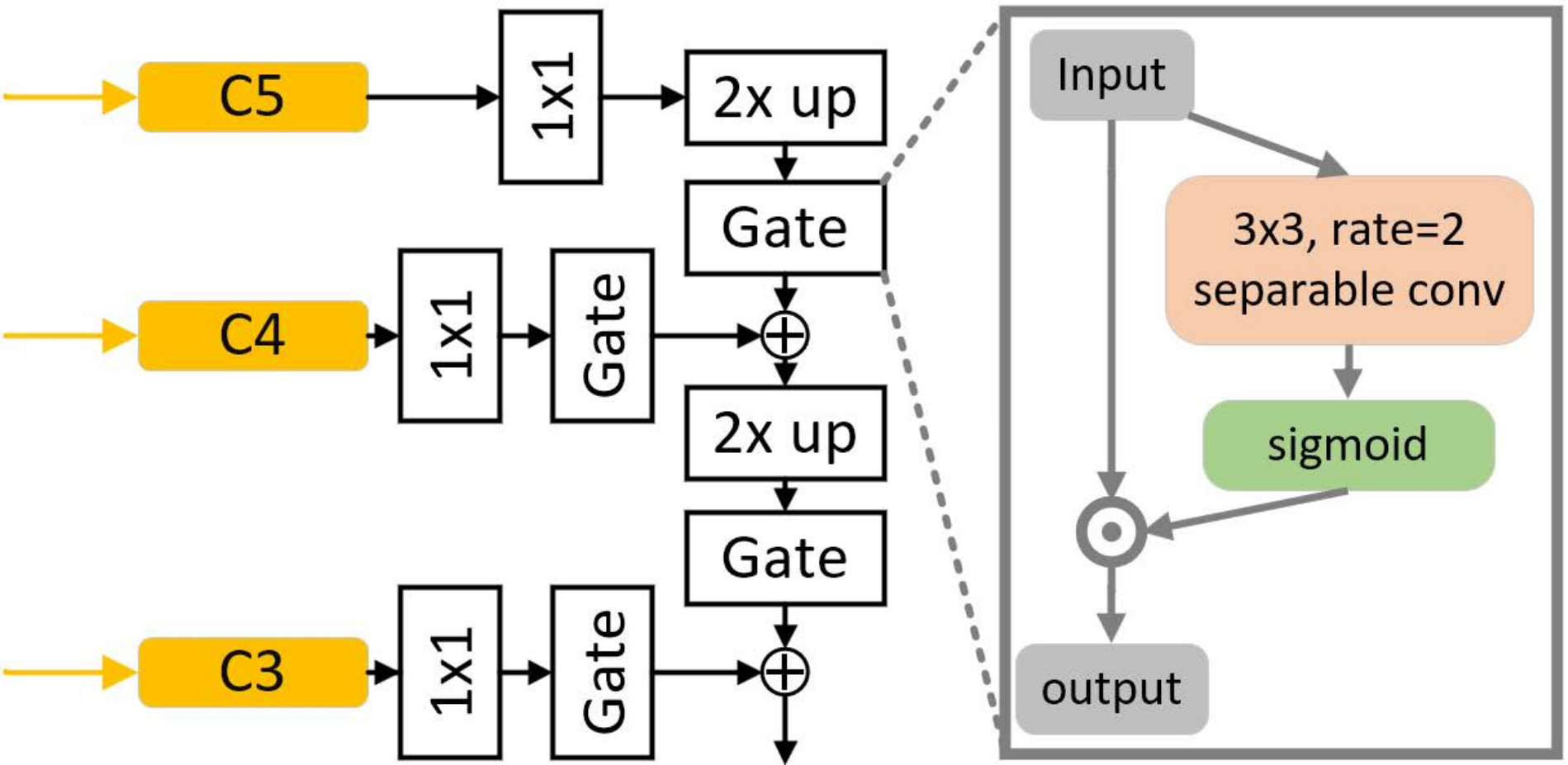}
	\caption{Gate mechanism in GFPN. Note that the separable convolution is shared by two \textcolor{black}{backbone} feature maps.}
	\label{fig:3}
\end{figure}

\textcolor{black}{To this end, we propose a flexible mechanism that enables the network to adaptively determine which backbone feature map should be the primary contributor in each level's feature learning so as to avoid the spatial shift and unexpected gradient propagation. Specifically, both backbone feature maps need to pass through a shared separable convolution layer, then the outputs are fed to a sigmoid activation respectively to form two score maps. Each score map has element-wise correspondence to the feature map with the element value in the scope of zero to one, therefore a score map can be regarded as a gate controlling the information flow between features and the proposed mechanism is named as Gate-FPN (GFPN). The gating mechanism is somewhat similar to the weighting method \cite{li2019, chen2019}. Specifically, we perform element-wise multiplication of two score maps with corresponding backbone feature maps, and add both products together to form a feature map in FPN. The calculation process can be denoted as formulas (1) through (3) where $x_1$ and $x_2$ denote the backbone feature maps, $w_s$ and $b_s$ denote the shared weights and biases respectively. Since the score maps are adaptively learned, we believe they play important role in preventing the spatial shift and unexpected gradient propagation. The GFPN mechanism is denoted as module `G' in Figure \ref{fig:2} and detailed as Figure \ref{fig:3}. Note that we share the same convolution in score map learning because we want to score each feature map under the same metric, and we prefer separable convolution instead of normal convolution because it reduces calculation and performs better. We use sigmoid activation because it works better than other typically used activation functions. The AP obtained by different activation functions can be seen in Table \ref{tab:handwash_given_pat}.}


\textcolor{black}{Each level feature map of GFPN will be sent to three branches}  --- classification, regression, and segmentation \textcolor{black}{that are represented by `class branch', `box branch' and `mask branch' in Figure \ref{fig:2}. ({\color{black} 3.1}) respectively}. Since the regression and classification branches are the same as those of RetinaNet, we ignore their details for simplicity \textcolor{black}{and discuss the mask branch in detail in the next subsection.} The three branches are processed in parallel. \textcolor{black}{Specifically,} we perform the same operations with shared parameters at each GFPN level and the output feature maps are used together to predict instances. The overall losses consist of three parts: classification, regression, and mask, \textcolor{black}{which are computed through the focal loss \cite{lin2018focal}, the ${\rm smoothed}$ $L_{1}$ loss \cite{ren2015faster} and the cross-entropy loss \cite{he2017mask} respectively}.
\begin{table}[!htbp]
	\caption{Comparison of different activation functions on Retina-FPN-50-400px}
	\centering
	\scriptsize
	\begin{tabular}{c|c|c|c|c|c}
		\toprule
	  \textbf{Gate} & $baseline$ & $Sigmoid$ & $Tanh$ & $Square$ & $R-T$\\ \midrule
	  \textbf{AP} & $30.5$ & $\textbf{31.1}$ & $30.2$ & $30.3$ & $30.8$\\%
	  \bottomrule
	\end{tabular}
	\label{tab:handwash_given_pat}
\end{table}

\vspace{5mm}
    \begin{spacing}{1.0}
    \par \small{\bfseries Forward}:
    \footnotesize\begin{equation}
        f_{1} \leftarrow x_{1} \cdot sigmoid(x_{1}w_{s}+b_{s});
    \end{equation}
    \footnotesize\begin{equation}
        f_{2} \leftarrow x_{2} \cdot sigmoid(x_{2}w_{s}+b_{s});
    \end{equation}
    \footnotesize\begin{equation}
        y \leftarrow f_{1} + f_{2};
    \end{equation}
    \par \small{\bfseries Back-propagation}:
    \footnotesize\begin{equation}
        L \leftarrow L(w_{s},b_{s};x_{1},y_{1},x_{2},y_{2})
    \end{equation}
    \footnotesize\begin{equation}
         f'_{1} \leftarrow \frac{dL}{df_{1}}
    \end{equation}
    \footnotesize\begin{equation}
        f'_{2} \leftarrow \frac{dL}{df_{2}}
   \end{equation}
   \footnotesize\begin{equation}
        w_{s} \leftarrow w_{s} - \alpha (f'_{1} \frac{\partial f_{1}}{\partial w_{s}} + f'_{2} \frac{\partial f_{2}}{\partial w_{s}})
    \end{equation}
    \footnotesize\begin{equation}
        b_{s} \leftarrow b_{s} - \alpha (f'_{2} \frac{\partial f_{2}}{\partial b_{s}} + f'_{2} \frac{\partial f_{2}}{\partial b_{s}})
    \end{equation}
    \footnotesize\begin{equation}
        x'_{1} \leftarrow \frac{\partial f_{1}}{\partial w_{s}}
    \end{equation}
    \footnotesize\begin{equation}
        x'_{2} \leftarrow \frac{\partial f_{2}}{\partial w_{s}}
    \end{equation}
    \par
    \end{spacing}

\subsection{Mask Branch}

\vspace{8pt}

To address the two problems \textcolor{black}{proposed in the beginning of Section \ref{SPRNet}}, we propose an Encoder-and-Decoder-like structure. The encoder part \textcolor{black}{is primarily referred to as the GFPN learning,} we \textcolor{black}{nevertheless propose} a multi-scale fusion \textcolor{black}{scheme to enhance the encoder part} so that details of object's morphological information are embedded into each pixel. After that, we sample pixels which are most likely to locate at the center of the instances, and perform a decoder \cite{DBLP:journals/corr/abs-1802-02611}-like reconstruction process.  \textcolor{black}{We name these pixels as positive pixels, each of which is gradually reconstructed as a 32 $\times$ 32 score map that denotes the mask of its corresponding instance.} See Figure \ref{fig:2}. ({\color{black} 3.2}) for more details.

\vspace{6pt}

\begin{figure}
	\centering
	\includegraphics[width=0.45\textwidth]{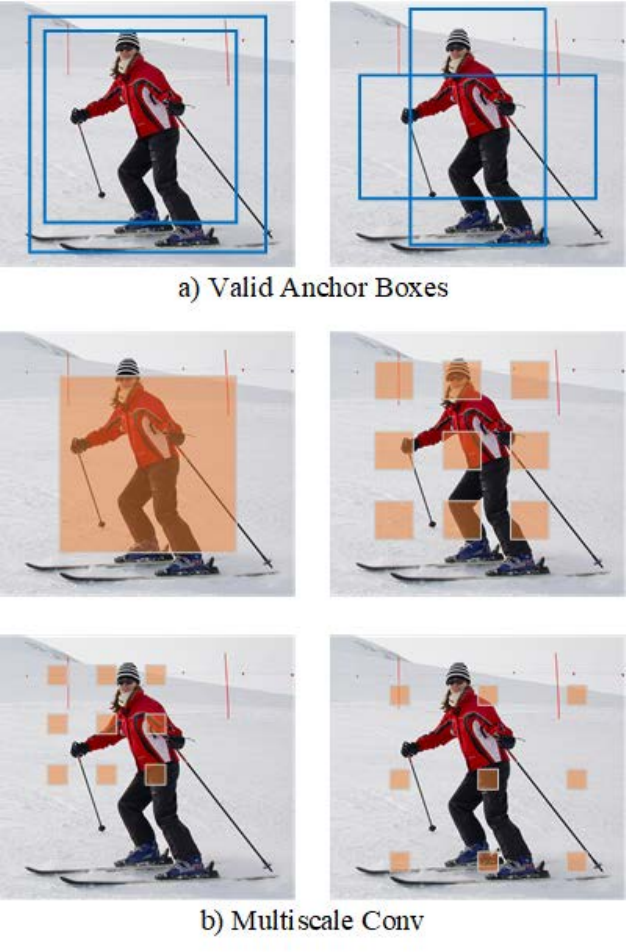}
	\caption{Multiscale convs could properly capture local and global information of instances. Kernels are designed to \textcolor{black}{have the same} size as any possible valid anchor boxes, so that the \textcolor{black}{positive pixel} will carry sufficient information.}
	\label{fig:32}
\end{figure}

\noindent \textbf{Multi-scale Fusion:} To endow a single pixel with enough information, \textcolor{black}{we apply four additional convolution operations to feature maps before mask estimation}, including a $1\times 1$ convolution with 256 channels and three $3\times 3$ convolutions with dilation rates at [2, 4, 6], and each has 128 channels, see Figure \ref{fig:32} for more details. After \textcolor{black}{channel-wise} concatenating these feature maps, each single pixel carries morphological information across various scales and has large receptive fields embedded.
\textcolor{black}{The usage of the convolutions with dilation is necessary because the sampled pixel may not locate at the very center of an instance, it therefore has to see `wider' for better capturing information of the entire instance.}

\vspace{4pt}

\par In order to retrieve \textcolor{black}{the positive pixels}, we first generate 9 anchor boxes on each pixel. If any one of the 9 boxes has an overlap larger than 0.7 (higher than 0.5, the box training positive threshold) with any instance, we set this pixel to be a positive sample for training and its label \textcolor{black}{will be used to attain the instance mask}.
\textcolor{black}{Hence, the pixels near the same instance have exactly the same training targets}.
More details are in Figure 6.
It is very clear that the higher the overlap threshold, the more probable the sampled pixel is the center of the instance, and vice versa. We need to leverage \textcolor{black}{how to determine} the best threshold. First, a low threshold usually results in more sampled pixels \textcolor{black}{locating} outside the instances, which makes it very hard to generate a preferable and accurate mask and makes the training difficult. In contrast, a high threshold makes the training relatively easy, but results in the \textcolor{black}{instability of the network}. This is simply because we sample pixels that correspond to the highest 100 classification scores \textcolor{black}{during inference}, \textcolor{black}{but} we cannot guarantee that these pixels are at the very center of the instances. Therefore, appropriately setting the overlap threshold becomes vital for training mask branch. In our ablation studies, we compare how different thresholds affect the final mAP. According to our results, a 0.7 threshold achieves robustness while retaining easy training.

\vspace{6pt}

\noindent \textbf{Single Pixel Reconstruction:}
A shared decoder \cite{DBLP:journals/corr/abs-1802-02611} is \textcolor{black}{exploited to realize the accurate mask computation}, which \textcolor{black}{uses a series of operations to gradually reconstruct the segmentation masks from individual positive pixels.} \textcolor{black}{Specifically}, we first use three consecutive deconvolution layers with no activation to gradually generate the $8\times 8$ instance mask \textcolor{black}{from a positive pixel}, then we use the deonvolution with ReLU activation twice to construct the final mask. We also add a nearest interpolation shortcut from the $8\times 8$ mask to the final classification layer. Note that \textcolor{black}{we do not use any activation like \emph{Relu} after each of the first three deconvolutions because each feature map is quite small, and \emph{Relu} may set many neurons to zero such that the important information contained in the feature maps is destroyed}. After recovering the $32\times 32$-pixel instance mask, \textcolor{black}{we convolve the mask with $K$ $1\times 1$ kernels and sample the highest channel value at each pixel location from the $32\times 32\times K$ mask map according to the classification branch score to obtain the instance mask. }In the loss part, we use binary cross-entropy function and only calculate the on-class mask loss of all $K$ classes.

\subsection{Implementation details}
\par One-stage instance segmentation is a brand new approach, and we illustrate its important details as follows.

\vspace{4pt}

\par
\begin{figure}[ht]
	\centering
	\includegraphics[width=0.4\textwidth]{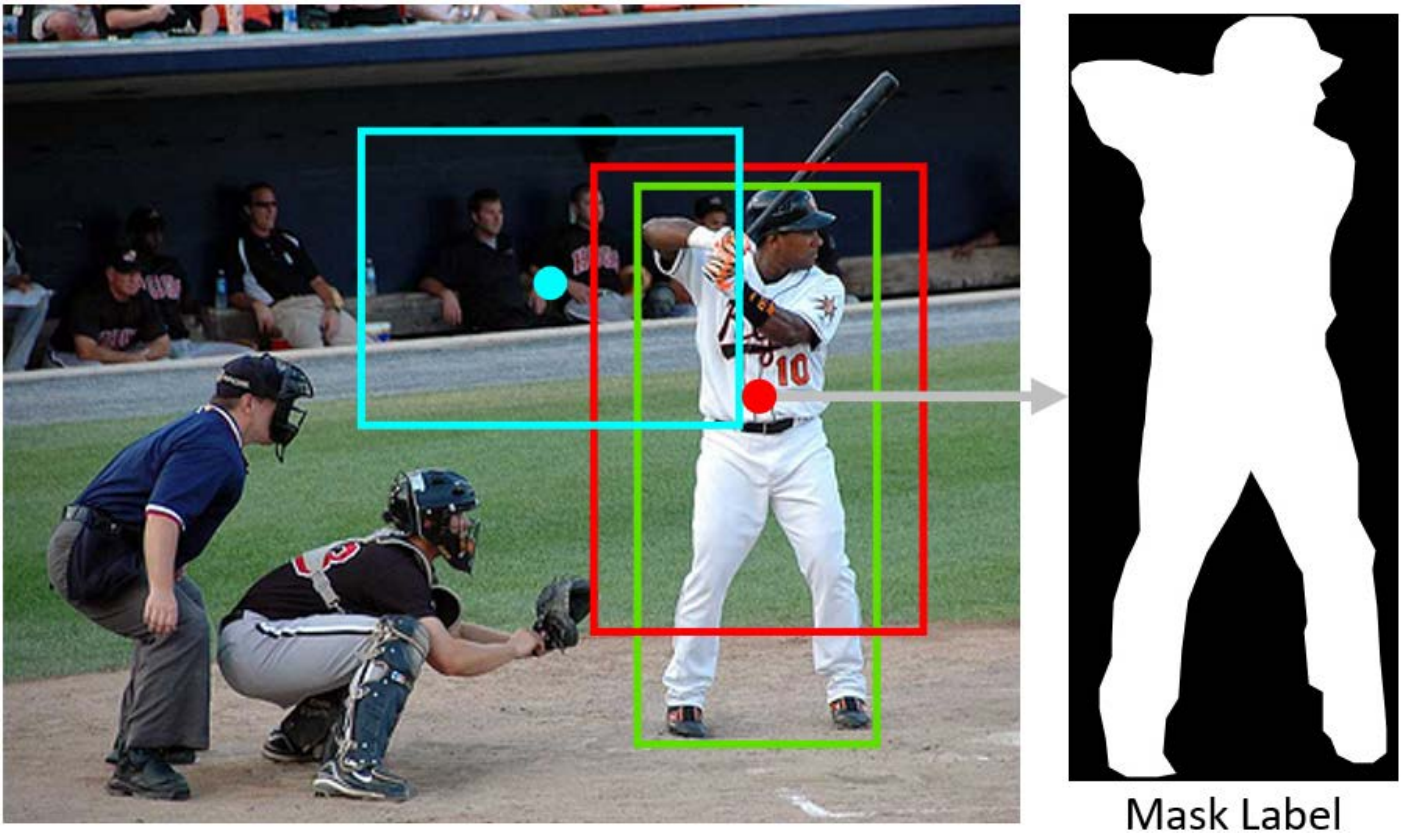}
	\caption{Green box is the target. Red box has an overlap greater than 0.7 with it, so the pixel at red dot is labeled as positive. Blue box is \textcolor{black}{hardly} overlapped with the target, hence, blue dot is labeled as negative.}
	\label{fig:4}
\end{figure}

 \noindent \textbf{Label Preparation:}
\textcolor{black}{We generate 9 anchor boxes with 3 different sizes and 3 different ratios on every pixel at each level of the feature maps. In box labeling, we set anchors that have overlaps larger than 0.5 with any instance label as the positive anchors, and set those anchors with overlaps less than 0.4 as the negative ones. In mask generation, we set the positive threshold to 0.7. The positive and negative anchors can be found in Figure \ref{fig:4}.}

\textcolor{black}{In the preparation of the training labels for mask prediction, we firstly find the anchors with overlaps larger than 0.7 with any target box at each pixel, and then select the anchor at each pixel location with highest overlap for training. We use 300 pixels corresponding to the 300 anchors, and each pixel is responsible for generating its corresponding instance.}

\par \noindent \textbf{Training:} SPRNet is easy to train. We train a total 25 epochs on the MS-COCO 2017 \texttt{train} dataset using Adam optimizer with initial learning rate of $10^{-5}$ and gradient clip at $10^{-3}$. We use ResNet backbone pre-trained on ImageNet, and train the entire network in an end-to-end way.

\begin{figure*}[ht]

	\centering
	
	\includegraphics[width=1.0\textwidth]{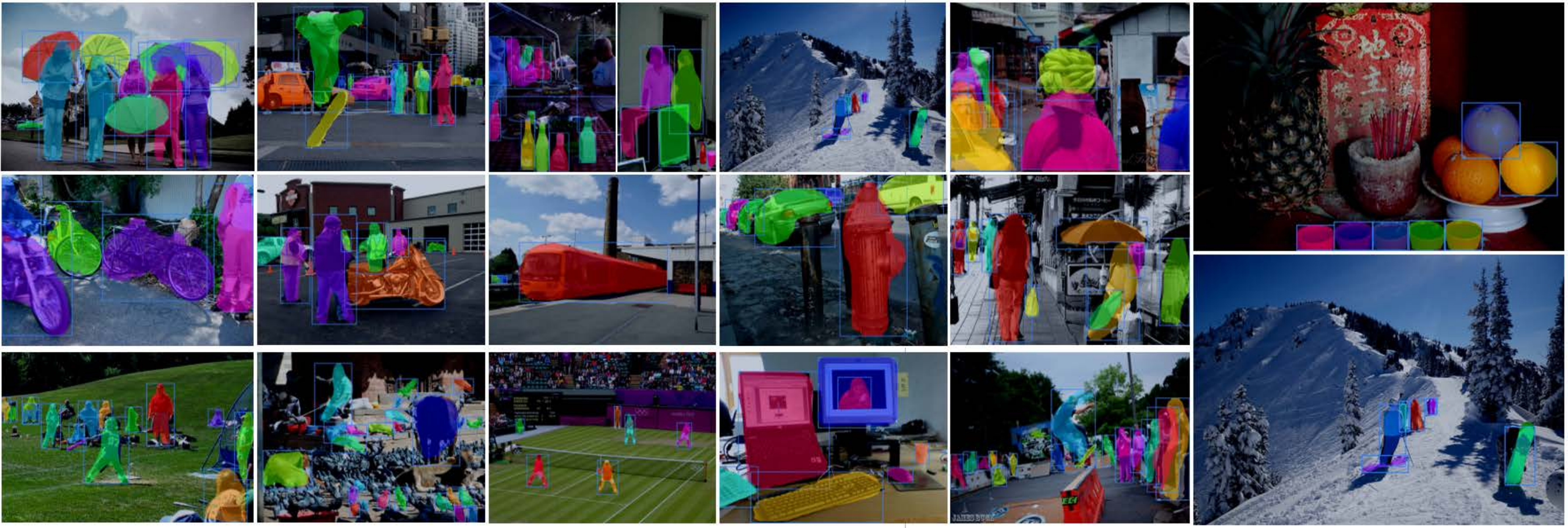}
	\caption{\textbf{Instance segmentation}: \textbf{SPRNet} generates boxes and masks with high accuracy and recall. Each instance is drawn in a distinguished color. }
	\label{fig:5}
\end{figure*}

\begin{table*}
	\normalsize
\caption{APs on MS-COCO 2017 \texttt{val} dataset, indicating that our model is not as powerful as Mask R-CNN but runs at a higher speed.}
	\centering
	\begin{tabular}{c|c|c|c|ccc|ccc}
		\Xhline{1.2pt}
										& \textrm{fps} & \textrm{scale} & \textrm{backbone}      & ${\rm AP}$ & ${\rm AP}_{50}$ & ${\rm AP}_{75}$ & ${\rm AP}_{S}$ & ${\rm AP}_{M}$ & ${\rm AP}_{L}$\\

		\cline{1-10}
		\textbf{Two-stage methods} &&&&&&&&&\\
		MNC \cite{dai2016instance}     &- & 800 & ResNet-101-C4         & 24.6 & 44.3      & 24.8      & 4.7      & 25.9     & 43.6      \\
		FCIS \cite{Yi2017Fully}+OHEM \cite{shrivastava2016training}    &- & 800 & ResNet-101-C5-dilated & 29.2 & 49.5      & -         & 7.1      & 31.3     & 50.0       \\
		FCIS +++\cite{Yi2017Fully}+OHEM \cite{shrivastava2016training} &- & 800 & ResNet-101-C5-dilated & 33.6 & 54.5      & -         & -        & -        & -      \\
		Mask R-CNN                    &7 & 800 & ResNet-50-FPN        & 33.6 & 55.2      & 35.3      & -    & -  & - \\
		Mask R-CNN                    &- & 800 & ResNet-101-C4        & 33.1 & 54.9     & 34.8     & 12.1    & 35.6   & 51.1 \\
		Mask R-CNN                    &5 & 800 & ResNet-101-FPN        & 35.7 & \textbf{58.0}      & 37.8      & 15.5    & 38.1   & 52.4 \\
		Mask R-CNN                    &5 & 800  & ResNet-101-GFPN        & \textbf{36.0} & 57.9      & \textbf{38.0}      & \textbf{16.0}    & \textbf{38.5}   & \textbf{54.0} \\
		\cline{1-10}
		\textbf{One-stage methods} &&&&&&&&&\\
		SPRNet (Ours)             & \textbf{10}  & 500 & ResNet-50-FPN  & 29.8 & 52.8 & 30.0 & 11.3   & 32.6       & 48.9      \\
		SPRNet (Ours)             & \textbf{10}  & 500 & ResNet-50-GFPN  & 30.4 & 53.2 & 30.8 & 12.4   & 33.3       & 49.6      \\
		SPRNet (Ours)             & \textbf{9}  & 800 & ResNet-50-GFPN  & 32.0 & 54.3 & 32.1 & 13.8   & 34.5       & 49.3      \\
		SPRNet             & \textbf{7}  & 800  & ResNet-101-GFPN  & \textbf{34.0} & \textbf{56.6} & \textbf{34.4} & \textbf{14.9}   & \textbf{35.4}       & \textbf{50.6}      \\
		\Xhline{1.2pt}		
	\end{tabular}
		\label{tab:1}
\end{table*}
\par \noindent \textbf{Inference:} During inference, the only difference \textcolor{black}{from training} is that we no longer use anchor overlap as the metric to sample pixels since there are no target labels, we use pixels with the highest 100 scores outputted by the classification branch, and after each sampled pixel is reconstructed into a $32\times 32$ instance mask, we use bilinear interpolation to resize them to the actual box size outputted by the regression branch.

\section{Experiments on Instance Segmentation}
\label{exp:seg}
In this section, we provide comparative experiments with the current state-of-the-art methods, along with comprehensive ablation experiments. All the experiments are conducted on MS-COCO dataset \cite{lin2014microsoft}. We report the standard MS-COCO metrics including AP, ${\rm AP}_{50}$, ${\rm AP}_{75}$ and ${\rm AP}_{S}$, ${\rm AP}_{M}$ and ${\rm AP}_{L}$ (\texttt{Average Precision} at different scales). We report the results in terms of box AP and mask AP jointly. Follow the previous work in \cite{he2017mask}, we train using 115k training images (\texttt{train}), and report ablations on the 5k validation images (\texttt{val}).
\begin{figure*}[ht]

	\centering
	\includegraphics[width=1.0\textwidth]{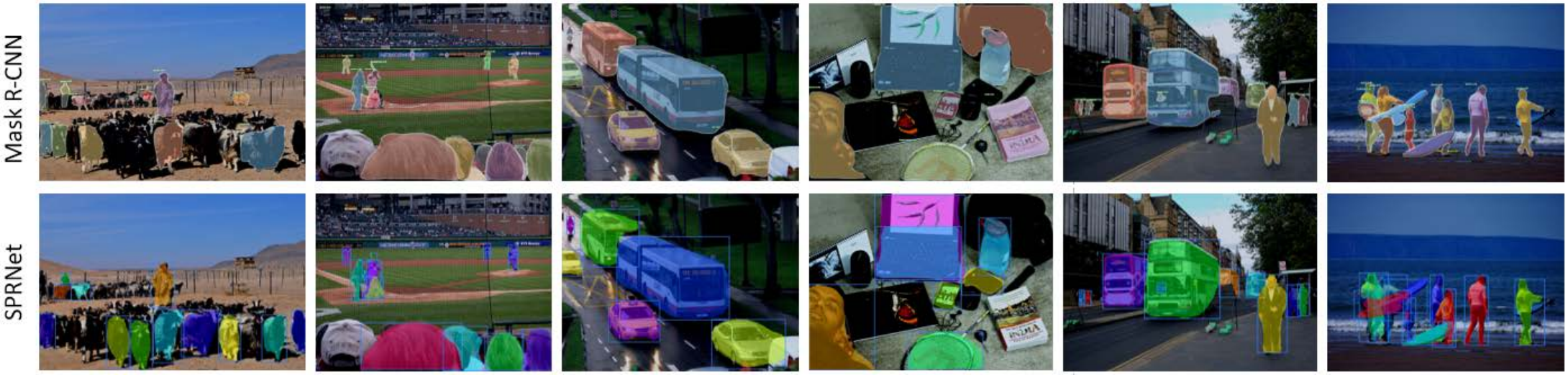}
	\caption{Mask R-CNN \textit{vs.} SPRNet. With the natural advantage of one-stage frameworks, SPRNet has a higher recall than Mask R-CNN (the first column), while Mask R-CNN has better performance in instance details.}
	\label{fig:6}
\end{figure*}

\begin{table*}
\caption{Ablation experiments concern the results of three main parts of SPRNet and threshold w.r.t. mask prediction. (a) \textbf{Fusion Paths:} obtain one pixel with large receptive fields (two strategies). (b) \textbf{FPN \textit{vs.} GFPN:} The structural feature learning (gated and non-gated). (c) \textbf{Mask Generation with Shortcuts:} the final mask generation with and without shortcuts. (d) \textbf{Mask IoU Threshold:} A proper threshold could maximize the performance.}
	\small
	\subtable[\textbf{Fusion Paths}]{
		\centering
		\begin{tabular}{l|p{0.68cm}p{0.68cm}p{0.68cm}|p{0.68cm}p{0.68cm}l}
			\Xhline{1.2pt}
							& ${\rm AP}$ & ${\rm AP}_{50}$ & ${\rm AP}_{75}$  & ${\rm AP}_{S}$ & ${\rm AP}_{M}$ & ${\rm AP}_{L}$ \\
			\hline
			C33x4          & 29.2    & 52.1   & 30.0   & 12.2   & 32.4       & 48.6     \\
			\hline
			C33-1,2,4,6  & 30.4    & 53.2   & 30.8   & 12.4   & 33.3       & 49.6        \\
            \Xhline{1.2pt}
		\end{tabular}
		\label{tab:2a}
	}
	\subtable[\textbf{FPN \textit{vs.} GFPN}]{
		\centering
		\begin{tabular}{l|p{0.68cm}p{0.68cm}p{0.68cm}|p{0.68cm}p{0.68cm}l}
\Xhline{1.2pt}
							& ${\rm AP}^{m}$ & ${\rm AP}^{m}_{50}$ & ${\rm AP}^{m}_{75}$ & ${\rm AP}^{bb}$ & ${\rm AP}^{bb}_{50}$ & ${\rm AP}^{bb}_{75}$ \\
			\hline
			FPN & 29.8  & 52.8  & 30.0  & 32.5  & 50.9    & 34.8            \\
			\hline
			GFPN & 30.4  & 53.2  & 30.8  & 33.6  & 52.0    & 35.9         \\
\Xhline{1.2pt}
		\end{tabular}
		\label{tab:2b}
	}
	\subtable[\textbf{Mask Generation with Shortcuts}]{
		\centering
		\begin{tabular}{l|p{0.54cm}p{0.54cm}p{0.54cm}|p{0.54cm}p{0.54cm}l}
\Xhline{1.2pt}
					& ${\rm AP}$ & ${\rm AP}_{50}$ & ${\rm AP}_{75}$  & ${\rm AP}_{S}$ & ${\rm AP}_{M}$ & ${\rm AP}_{L}$\\
			\hline
			Deconv                & 30.0   & 52.9 & 30.5   & 12.4   & 32.6       & 49.2       \\
			\hline
			Deconv+shortcut     & 30.4    & 53.2   & 30.8   & 12.4   & 33.3       & 49.6   \\
\Xhline{1.2pt}
		\end{tabular}
		\label{tab:2c}
	}
	\quad
	\quad
	\subtable[\textbf{Mask IoU Threshold}]{
		\centering
		\begin{tabular}{l|p{0.65cm}p{0.65cm}p{0.65cm}|p{0.65cm}p{0.65cm}l}
\Xhline{1.2pt}
			Overlap     & ${\rm AP}$ & ${\rm AP}_{50}$ & ${\rm AP}_{75}$  & ${\rm AP}_{S}$ & ${\rm AP}_{M}$ & ${\rm AP}_{L}$\\
			\hline
			$>0.5 $   & 30.2   & 53.0    & 30.6  & 12.1   & 33.3       & 49.3     \\
			\hline
			$>0.7 $   & 30.4    & 53.2   & 30.8   & 12.4   & 33.3       & 49.6    \\
\Xhline{1.2pt}
		\end{tabular}
		\label{tab:2d}
	}
	
	\label{tab:2}
\end{table*}

\subsection{Main Results}

We compare SPRNet with the state-of-the-art instance segmentation methods in Table \ref{tab:1}. All instantiations of our model have comparable performance with previous state-of-the-art models including MNC \cite{dai2016instance}, FCIS \cite{Yi2017Fully}, Mask R-CNN \cite{he2017mask}, and the winners of MS-COCO 2015, 2016 and 2017 segmentation challenges. Without embellishments or any data augmentation, \textcolor{black}{the AP of the SPRNet with ResNet-50-GFPN backbone is 1.6 percents lower than that of the Mask R-CNN with ResNet-50-FPN that also includes horizontal flip training and online hard example mining (OHEM) \cite{shrivastava2016training}. However, the SPRNet with ResNet-50-GFPN backbone has higher computational efficiency than the comparative method owing to the one-stage framework.
In addition, we obtain higher detection recall than R-CNN based frameworks due to the natural advantage of one-stage detectors. More importantly,} our method could robustly distinguish edges on overlapped objects. The overlapping issue is extremely severe in FCIS and later solved by Mask R-CNN. Our one-stage SPRNet has shown good capability of handling this issue.
We visualize some examples of the SPRNet predictions in Figure \ref{fig:5}. SPRNet achieves good results even under challenging conditions. Examples contain crowded objects, objects in extreme sizes, objects with uncommon morphology. Our methods show promising results in distinguishing borders and maintaining edge consistence.

\subsection{Ablation Experiments}
To better understand the reason of SPRNet's effectiveness, we run a number of ablations. Results are shown in Table \ref{tab:2} and discussed in detail below.

\noindent \textbf{Fusion Paths:} In Table \ref{tab:2a}, we show two alternative strategies to obtain one pixel with large receptive fields. \textbf{C33x4} means four \emph{consecutive} 3$\times$3 convolutions, which is as same as the classification or the regression branch; \textbf{C33-1,2,4,6} means four \emph{parallel} 3$\times$3 convolutions with different dilation rates [1,2,4,6], which has similar computation costs as C33x4, but is able to capture more morphological transformation of an instance. Results show 1.2 percents' improvement in AP. Moreover, with the increasing of object shapes, their performance gap gets larger.

\vspace{4pt}

\par \noindent \textbf{FPN \textit{vs.} GFPN:} Table \ref{tab:2b} ablates the performance gap between FPN and the proposed GFPN. We provide the comparative results about mask mAP and box mAP. When evaluating box predictions, we remove the mask branch from SPRNet to make it degrade to a RetinaNet variant with GFPN. Results show GFPN outperforms FPN steadily under all criteria.

\vspace{4pt}

\noindent \textbf{Mask Generation with Shortcuts:} Table \ref{tab:2c} compares the methods w/ or w/o shortcut connection. For the Deconv+shortcut model, we add an additional shortcut from $8\times 8$ feature map (with 4$\times$ upsampling) to the final classification layer, and it improves the AP score by 0.4 percent.

\vspace{4pt}

\noindent \textbf{Mask IoU Threshold:} Table \ref{tab:2d} compares the models trained with different mask IoU threshold when preparing training samples. A small threshold (e.g., 0.5) results in an inferior AP score. \textcolor{black}{This is actually predictable based on the fact that a small threshold makes it hard to converge for training.}

\subsection{Analysis}
We compare SPRNet and Mask R-CNN in Figure \ref{fig:6}, with each instance drawn in a distinguished color. Both the methods use the same ResNet-50 backbone. The samples include objects at various scales. Our method shows comparable performance with Mask R-CNN in most cases. Specifically, Mask R-CNN has a higher accuracy (i.e., more accurate mask boundaries for the detected objects) while SPRNet has a higher recall (i.e., more objects are detected). Mask R-CNN beats SPRNet in extreme situations with more detailed and aligned predictions. \textcolor{black}{We deem the performance difference predictable because Mask R-CNN uses its final box predictions to accurately sample RoIs, making mask prediction a very easy binary semantic segmentation task. In contrast, one-stage frameworks must find another totally different but plausible approach to achieve accurate generation of the instance masks while improving the computational speed. Despite using atrous convolution to compact spatial information into a single pixel, it is still challenging to recover a very detailed and accurate mask from pixels. Also, due to the one-stage framework, regression values will largely influent the final accuracy in both box and mask prediction. For regression boxes that are seriously shifted from the ground truth boxes, mask could easily be mis-aligned. However, SPRNet delivers a higher speed. Under the same experimental settings (i.e., GTX 1080Ti), Mask R-CNN using ResNet-50-FPN as the backbone runs at 7 fps, while our SPRNet runs at 9 fps, which is about 30\% faster. }

\section{Additional Experiments}
\label{add:exp}
Besides mask AP evaluation metric, we provide two more sets of experiments, concerning the AP of the BBox (bounding box) object detection and \texttt{Average Recall (AR)} of both object detection and instance segmentation. We justify the effect of GFPN on box detection in the first set of experiments, in which we find that SPRNet could outperform RetinaNet, especially in objection detection with more detailed categories. It's noteworthy that as the input size increases, RetinaNet shows a downtrend in large objects detection, while SPRNet effectively alleviates this problem, and this is because the GFPN prevents the gradients from passing from lower levels to higher ones, making the FPN-like structure performs as similar as the ordinary C4 or C5-based frameworks in large objects detection. In the second set of experiments, we compare the AR results of both box detection and mask prediction using RetinaNet, Mask R-CNN and SPRNet respectively, and find that SPRNet has largely increased the overall number of detected objects with small, median and large sizes.
\vspace{4pt}

\subsection{BBox Object Detection}
\label{exp:bbox}

\begin{table*}
	\scriptsize
	\caption{\textbf{Object Detection}: box AP on MS-COCO. The comparison between famous one and two-stage frameworks shows that using the same backbone, SPRNet with GFPN has surpassed every existing one-stage detectors with a remarkable improvement on small object detection accuracy.}
	\centering
	\begin{tabular}{c|c|ccc|ccc}
\Xhline{1.2pt}
														& \textrm{backbone}                                     & ${\rm AP}^{bb}$ & ${\rm AP}^{bb}_{50}$ & ${\rm AP}^{bb}_{75}$ & ${\rm AP}^{bb}_{S}$ & ${\rm AP}^{bb}_{M}$ & ${\rm AP}^{bb}_{L}$ \\

		\hline
		\textbf{Two-stage methods} &&&&&&& \\
		Faster R-CNN+++                                & ResNet-101-C4                                  & 34.9      & 55.7           & 37.4           & 15.6          & 38.7          & 50.9              \\
		Faster R-CNN w FPN                             & ResNet-101-FPN                                 & 36.2      & 59.1           & 39.0           & 18.2          & 39.0          & 48.2              \\
		Faster R-CNN by G-RMI \cite{huang2017speed}     & Inception-ResNet-v2 \cite{szegedy2017inception} & 34.7      & 55.5           & 36.7           & 13.5          & 38.1          & 52.0            \\
		Faster R-CNN w TDM \cite{Shrivastava2016Beyond} & Inception-ResNet-v2-TDM                        & 36.8      & 57.7           & 39.2           & 16.2          & 39.8          & 52.1            \\
		Faster R-CNN, RoIAlign                         & ResNet-101-FPN                                 & 37.3      & 59.6           & 40.3           & 19.8          & 40.2          & 48.8            \\
		Mask R-CNN                                     & ResNet-101-FPN                                 & 38.2      & 60.3           & 41.7           & 20.1          & 41.1          & 50.2           \\
		\hline
		\textbf{One-stage methods} &&&&&&& \\
		YOLO v2 \cite{Redmon2017YOLO9000}                                        & Darknet-19                                     & 21.6      & 44.0           & 19.2           & 5.0           & 22.4          & 35.5              \\
		YOLO v3 608x608                                & Darknet-53                                     & 33.0      & 57.9           & 34.4           & 18.3          & 35.4          & 41.9           \\
		DSSD513                                        & ResNet-101-DSSD                                & 33.2      & 53.3           & 35.2           & 13.0          & 35.4          & \textbf{51.1}              \\
		RetinaNet 500       & ResNet-50-FPN & 32.5 & 50.9      & 34.8    & 13.9    & 35.8 & 46.7       \\
		RetinaNet 800       & ResNet-50-FPN & 35.7 & 55.0    & 38.5    & 18.9    & 38.9       & 46.3       \\
		SPRNet (Ours) 500    & ResNet-50-GFPN & 33.6 & 52.1 & 35.9  & 15.1  & 37.4  & 49.7       \\
		SPRNet (Ours) 800   & ResNet-50-GFPN  & \textbf{36.0} & \textbf{55.3}  & \textbf{38.8}  & \textbf{20.2}    & \textbf{39.6}   & 48.6              \\
\Xhline{1.2pt}
	\end{tabular}

	\label{tab:3}

\end{table*}

\begin{table*}
	\scriptsize
	\centering
	\caption{\textbf{Object Detection:} box AP of RetinaNet with FPN or GFPN. We use ResNet-50 as the common backbone, and we compare detailed AP under different input sizes. We report box AP without training or using the mask branch, which means we directly compare FPN and GFPN in terms of detection. } \begin{tabular}{p{0.9cm}<{\centering}|p{1.1cm}<{\centering}|p{0.9cm}<{\centering}p{0.9cm}<{\centering}p{0.9cm}<{\centering}|p{0.9cm}<{\centering}p{0.9cm}<{\centering}p{0.9cm}<{\centering}}
		\Xhline{1.2pt}
\textrm{scale}  &GFPN?& ${\rm AP}^{bb}$ & ${\rm AP}^{bb}_{50}$ & ${\rm AP}^{bb}_{75}$ & ${\rm AP}^{bb}_{S}$ & ${\rm AP}^{bb}_{M}$ & ${\rm AP}^{bb}_{L}$\\

		\cline{1-8}
		\multirow{2}{*}{400}  &  & 30.5 & \textbf{47.8} & 32.7  & 11.2  & 33.8 & 46.1 \\
		&\checkmark & \textbf{31.1}  &47.3 & \textbf{32.9}  & \textbf{11.5}  &\textbf{34.9} & \textbf{50.4}     \\
		\cline{1-8}
		\multirow{2}{*}{500}   &  & 32.5      & 50.9   & 34.8      & 13.9          & 35.8          & 46.7         \\
		 &\checkmark  & \textbf{33.6}      & \textbf{52.0}        & \textbf{35.9}   & \textbf{15.0}          & \textbf{37.4}          & \textbf{49.8}      \\
		\cline{1-8}
		\multirow{2}{*}{600}   &  & 34.3   & 53.2     & 36.9     & 16.2          & 37.4          & 47.4    \\
		&\checkmark & \textbf{35.0}      & \textbf{54.0}       & \textbf{37.5}   & \textbf{17.8}          & \textbf{39.0}          & \textbf{49.9}     \\
		\cline{1-8}
		\multirow{2}{*}{700}  & & 35.1      & 54.2   & 37.7     & 18.0     & 39.3    & 46.4   \\
		 &\checkmark  & \textbf{35.6}      & \textbf{54.7}   & \textbf{38.0}   & \textbf{18.6}  & \textbf{39.8}       & \textbf{48.9}    \\
		\cline{1-8}
		\multirow{2}{*}{800}   & & 35.7      & 55.0     & 38.5   & 18.9    & 38.9     & 46.3   \\
		&\checkmark & \textbf{36.0}    & \textbf{55.2}  & \textbf{38.8}  & \textbf{20.1}    & \textbf{39.6}   & \textbf{48.6}   \\
	\Xhline{1.2pt}	
	\end{tabular}
		\label{tab:4}
\end{table*}

We compare SPRNet \textcolor{black}{with} the state-of-the-art MS-COCO bounding-box object detection in Table \ref{tab:3}. With an input size of 500 pixels, we \textcolor{black}{achieve 1.1 percents' improvement in box AP} compared with best one-stage detector RetinaNet.

To analyze the effect of the gating mechanism for FPN, we conduct experiments to compare RetinaNet with GFPN or traditional FPN \textcolor{black}{and the results are demonstrated} in Table \ref{tab:4}. By introducing the gate mechanism, the RetinaNet with GFPN outperforms the other one with FPN by 1.2, 1.6 and 3.1 percents' in AP for small, medium and large object detection, respectively. The effectiveness of gradient blocking is also shown in Table \ref{tab:4}, with an improvement of up to 4.3 percents' in AP for large object detection. Similarly, the effectiveness of the gating mechanism is well shown \textcolor{black}{by the AP improvement of up to 1.8 percents' for small object.}

Using ResNet-50 as the backbone, SPRNet easily beats all existing one-stage methods. Our elaborately designed gating mechanism has shown promising effects in solving previously mentioned problems of the common `bottom up' structure. By automatically abandoning inferior information between features, we find a promising approach to improve the performance of detecting small objects. By blocking gradient flows, larger objects detection becomes isolated to smaller ones, which yields more than 2 \textcolor{black}{percents'} improvement in final AP. Especially, it is the first time that a one-stage method \textcolor{black}{using ResNet-50} beats the best two-stage methods using ResNet-101 and RoIAlign at amazing 20.2 \textcolor{black}{percents in ${\rm AP}_{S}$}.

\begin{table*}
	\centering
\caption{(test on val) Box ARs of RetinaNet and Mask-RCNN with or without GFPN. Significant improvement on ARs can be observed.}
    \normalsize
	\begin{tabular}{c|c|c|c|c|c|c|c}
\Xhline{1.2pt}
										& \textrm{backbone} & ${\rm AR}_{1}^{\rm box}$ & ${\rm AR}_{10}^{\rm box}$ & ${\rm AR}_{100}^{\rm box}$ & ${\rm AR}_{S}^{\rm box}$ & ${\rm AR}_{M}^{\rm box}$ & ${\rm AR}_{L}^{\rm box}$\\

		\cline{1-8}
		
        Mask R-CNN  & ResNet-101-FPN & 32.0 & 50.1      & 52.4      & 32.3   & 56.7   & 67.5 \\
        Mask R-CNN  & ResNet-101-GFPN & \textbf{32.9} & \textbf{51.2}      & \textbf{53.5}      & \textbf{32.8}    & \textbf{57.5}   & \textbf{69.0} \\
		\cline{1-8}
        RetinaNet & ResNet-50-FPN  & 30.7 & 49.1 & 52.0 & 32.0  & 56.9       & 68.0   \\
		RetinaNet & ResNet-50-GFPN  & 31.6 & 50.0 & 53.1 & 34.4   &57.6       & 68.6     \\
		RetinaNet & ResNet-101-FPN  & 32.2 & 50.8 & 53.8 & 34.1  & 58.4       & 70.5    \\
		RetinaNet & ResNet-101-GFPN  & \textbf{33.1} & \textbf{51.8} & \textbf{54.9} & \textbf{36.6}   & \textbf{59.2}       & \textbf{71.0}      \\
\Xhline{1.2pt}
		
	\end{tabular}
	
	\label{tab:22}
\end{table*}

\begin{table*}
	\centering
	\caption{(test on val) Segmentation ARs of SPRNet and Mask-RCNN with or without GFPN. Significant improvement on ARs can be observed.}
    \normalsize
	\begin{tabular}{c|c|c|c|c|c|c|c}
\Xhline{1.2pt}
										& \textrm{backbone} & ${\rm AR}_{1}^{\rm seg}$ & ${\rm AR}_{10}^{\rm seg}$ & ${\rm AR}_{100}^{\rm seg}$ & ${\rm AR}_{S}^{\rm seg}$ & ${\rm AR}_{M}^{\rm seg}$ & ${\rm AR}_{L}^{\rm seg}$\\

		\cline{1-8}
        Mask R-CNN  & ResNet-101-FPN & 29.9 & 45.7      & 47.6      & 26.2   & 51.8   & 65.1 \\
        Mask R-CNN  & ResNet-101-GFPN & \textbf{30.3} & \textbf{45.9}      & \textbf{47.7}      & \textbf{26.3}    & \textbf{51.7}   & \textbf{64.7} \\
		\cline{1-8}
		SPRNet & ResNet-50-FPN  & 28.6 & 44.6 & 46.9 & 26.0   & 52.0       & 65.1 \\
        SPRNet & ResNet-50-GFPN  & 29.5 & 45.4 & 47.9 & 27.9   & 52.7       & 65.6 \\

        SPRNet & ResNet-101-FPN  & 30.0 & 46.1 & 48.5 & 27.7   & 53.4       & 67.5 \\
		SPRNet & ResNet-101-GFPN  & \textbf{30.9} & \textbf{47.0} & \textbf{49.5} & \textbf{29.7}   & \textbf{54.1}       & \textbf{68.0} \\
\Xhline{1.2pt}
		
	\end{tabular}

	\label{tab:23}
\end{table*}
\par In Table \ref{tab:4}, we compare only the effect of GFPN by removing the mask branch both in training and inference. It is noteworthy that \textcolor{black}{the RetinaNet's detection accuracy of large objects} is far less than our methods. As we mentioned before, GFPN \textcolor{black}{provides} a better way \textcolor{black}{to} feature fusion, and this is the reason of improvements on small object detection. In the meanwhile, GFPN restricts the gradient propagation from low levels into higher levels, and this is the reason of improvements on large object detection. These improvements \textcolor{black}{happen} on every input scale. More importantly, on the metric of ${\rm AP}_{50}$, GFPN has \textcolor{black}{been} improved by up to \textcolor{black}{1.1 percents in AP}, which means more objects are likely to be well detected. On the metric of ${\rm AP}_{75}$, GFPN has \textcolor{black}{outperformed} FPN by up to \textcolor{black}{1.1 percents in AP}, and \textcolor{black}{this is very important} for practical application \textcolor{black}{because} our methods delivers more accurate detection on objects that are most likely to be detected. We do not try other input scale that is smaller than 400 or larger than 800 pixels, the reason is that existing ResNet backbone may fail in extracting spatial information when objects get too small, and convolution kernels may also be limited in receptive fields when objects get too large.

\subsection{Recall on objects}
\label{exp:recall}

One of the most important metrics to evaluate the detection performance is the \textit{Average Recall}. Given an image, the more instances are detected, the higher AR it will get.

\par For two-stage detectors, recalls could be considered in two parts. The first is \textcolor{black}{F}PN's recall, where anchors will be classified as positive or negative, and only positive boxes will be sent into following subnetworks, where the final recall is derived. While in one-stage frameworks, without proposal network, we will only get one recall.

\par In Table \ref{tab:22} and Table \ref{tab:23}, we compare Mask R-CNN, RetinaNet and SPRNet in both object detection and instance segmentation tasks.

Implemented on Mask R-CNN and RetinaNet, Gate-FPN has largely increased the AR under all metrics.
\textcolor{black}{By simply importing Gate-FPN, SPRNet has not only surpassed Mask R-CNN in overall AR, but also achieved prominent improvements on small object detection. The results prove that Gate-FPN is a very solid and flexible module for general use in different detection frameworks.}

\section{Conclusions}
\label{conc}
\textcolor{black}{We present the SPRNet as a one-stage approach to image instance segmentation}, without introducing the region proposals. SPRNet achieves comparable performance with the state-of-the-art two-stage models while running at a faster speed. By introducing GFPN, we bring one-stage detectors into a higher level \textcolor{black}{in} that \textcolor{black}{we enable it} to deliver better detection than prevalent one- and two-stage detectors. This work represents a feasible solution for delivering accurate and fast instance-level recognition.

\ifCLASSOPTIONcaptionsoff
  \newpage
\fi

\bibliography{references}{}
\bibliographystyle{IEEEtran}

\begin{IEEEbiography}[{\includegraphics[width=1in,height=1.25in,clip,keepaspectratio]{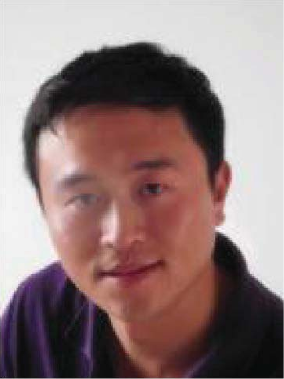}}]{Jun Yu} (M'13) received the B.Eng. and Ph.D. degrees from Zhejiang University, Zhejiang, China. He is currently a Professor with the School of Computer Science and Technology, Hangzhou Dianzi University, Hangzhou, China. He was an Associate Professor with the School of Information Science and Technology, Xiamen University, Xiamen, China. From 2009 to 2011, he was with Nanyang Technological University, Singapore. From 2012 to 2013, he was a Visiting Researcher at Microsoft Research Asia (MSRA). Over the past years, his research interests have included multimedia analysis, machine learning, and image processing. He has authored or coauthored more than 80 scientific articles. Prof. Yu serves as an associate editor for the IEEE Transactions on Circuits and Systems for Video Technology, Journal of Pattern Recognition, Journal of Information Sciences. He served as a program committee member or reviewer of top conferences and prestigious journals. He is a Professional Member of the Association for Computing Machinery (ACM) and the China Computer Federation (CCF).
\end{IEEEbiography}

\begin{IEEEbiography}[{\includegraphics[width=1in,height=1.25in,clip,keepaspectratio]{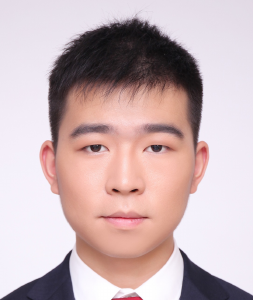}}]{Jinghan Yao} is an undergraduate of Hangzhou Dianzi University, Zhejiang, China. He is currently an senior student in the School of Honor, majors in Computer Science and Technology. Since 2015, he has been doing research on machine learning, image processing, computer vision and high performance computing.
\end{IEEEbiography}

\begin{IEEEbiography}[{\includegraphics[width=1in,height=1.25in,clip,keepaspectratio]{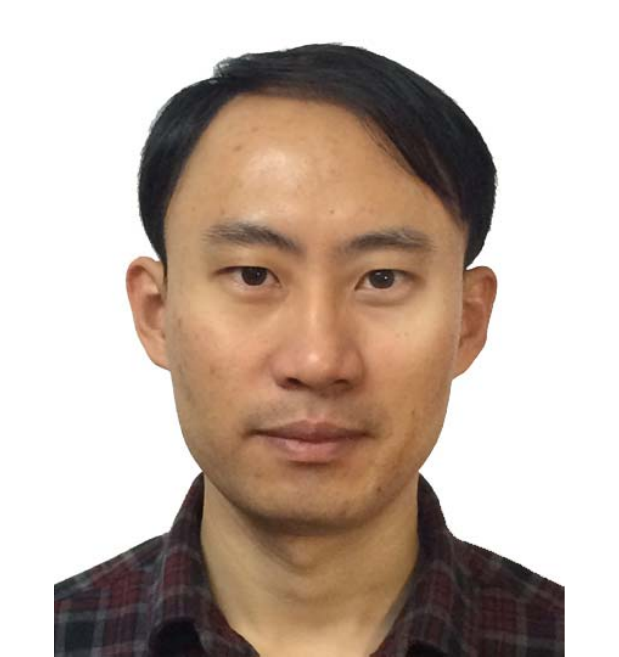}}]{Jian Zhang} received the Ph.D. degree from Zhejiang University, Zhejiang, China. He is currently an Associate Professor with the School of Science and Technology, Zhejiang International Studies University, Hangzhou, China. From 2009 to 2011, he was with Department of Mathematics of Zhejiang university as a Post-doctoral Research Fellow. In 2016, he had been doing research on machine learning at Simon Fraser University (SFU) as a Visiting Scholar. His research interests include but not limited to machine learning, computer animation and image processing. He was awarded the certificates of outstanding contribution in reviewing by Journal of Pattern Recognition and Neurocomputing.
\end{IEEEbiography}

\begin{IEEEbiography}[{\includegraphics[width=1in,height=1.25in,clip,keepaspectratio]{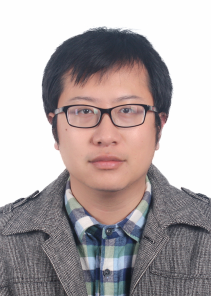}}]{Zhou Yu} received the B.Eng. and Ph.D. degrees from Zhejiang University, Zhejiang, China, in 2010 and 2015, respectively. He is currently a Lecturer with the School of Computer Science and Technology, Hangzhou Dianzi University, and his research interests includes multimodal data analysis, computer vision, machine learning and deep learning.
\end{IEEEbiography}

\begin{IEEEbiography}[{\includegraphics[width=1in,height=1.25in,clip,keepaspectratio]{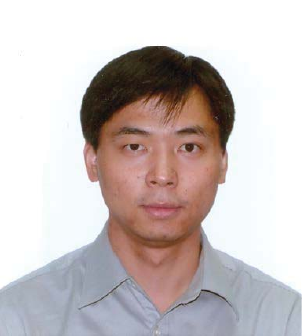}}]{Dacheng Tao} (F’15) is Professor of Computer Science and ARC Laureate Fellow in the School of Computer Science and the Faculty of Engineering and Information Technologies, and the Inaugural Director of the UBTECH Sydney Artificial Intelligence Centre, at the University of Sydney. He mainly applies statistics and mathematics to Artificial Intelligence and Data Science. His research results have expounded in one monograph and 200+ publications at prestigious journals and prominent conferences, such as IEEE T-PAMI, T-IP, T-NNLS,T-CYB, IJCV, JMLR, NIPS, ICML, CVPR, ICCV, ECCV, ICDM; and ACM SIGKDD, with several best paper awards, such as the best theory/algorithm paper runner up award in IEEE ICDM’07, the best student paper award in IEEE ICDM’13, the 2014 ICDM 10-year highest-impact paper award, the 2017 IEEE Signal Processing Society Best Paper Award, and the distinguished paper award in the 2018 IJCAI. He received the 2015 Austrlian Scopus-Eureka Prize and the 2018 IEEE ICDM Research Contributions Award. He is a Fellow of the Australian Academy of Science, AAAS, IEEE, IAPR, OSA and SPIE.
\end{IEEEbiography}

\end{document}